\documentclass[10pt,twocolumn,letterpaper]{article}

\usepackage{cvpr}              
\usepackage{color}
\usepackage{pifont}
\usepackage[accsupp]{axessibility}  

\newcommand{\method}{{JiSAM}}

\newcommand{\paragraphskip}{\vspace{0.15cm}}
\newcommand{\R}{\mathbb{R}}
\definecolor{cvprblue}{rgb}{0.21,0.49,0.74}
\usepackage[pagebackref,breaklinks,colorlinks,allcolors=cvprblue]{hyperref}
\definecolor{forest}{RGB}{34,139,34}

\def\paperID{12712} 
\def\confName{CVPR}
\def\confYear{2025}

\title{\method : Alleviate Labeling Burden and Corner Case Problems in Autonomous Driving via Minimal Real-World Data}

\author{
Runjian Chen$^{1}$\quad 
Wenqi Shao$^{2}$\footnotemark[1] \quad 
Bo Zhang$^2$\quad  
Shaoshuai Shi$^{3}$\quad 
Li Jiang$^{4}$\quad 
Ping Luo$^{1,5}$\footnotemark[1]\thanks{Corresponding authors.}\\[3mm]
$^1$The University of Hong Kong \quad
$^2$Shanghai AI Laboratory \\
$^3$Voyager Research, Didi Chuxing \quad
$^4$The Chinese University of Hong Kong (Shenzhen) \\
$^5$HKU Shanghai Intelligent Computing Research Center
\\
{\tt\small \{rjchen, pluo\}@cs.hku.hk  \quad shaowenqi@pjlab.org.cn}
}

\begin{document}
\maketitle
\begin{abstract}
Deep-learning-based autonomous driving (AD) perception introduces a promising picture for safe and environment-friendly transportation. However, the over-reliance on real labeled data in LiDAR perception limits the scale of on-road attempts. 3D real world data is notoriously \textbf{time-and-energy-consuming} to annotate and lacks \textbf{corner cases} like rare traffic participants. On the contrary, in simulators like CARLA, generating labeled LiDAR point clouds with corner cases is a piece of cake. However, introducing synthetic point clouds to improve real perception is non-trivial. This stems from two challenges: 1) sample efficiency of simulation datasets 2) simulation-to-real gaps. To overcome both challenges, we propose a plug-and-play method called \method\ , shorthand for  \textbf{Ji}ttering augmentation, domain-aware backbone and memory-based \textbf{S}ectorized \textbf{A}lign\textbf{M}ent. In extensive experiments conducted on the famous AD dataset NuScenes, we demonstrate that, with SOTA 3D object detector, \method\ \ is able to utilize the simulation data and only labels on $2.5\%$ available real data to achieve comparable performance to models trained on all real data. Additionally, \method\ \ achieves more than 15 mAPs on the objects not labeled in the real training set. Codes and models are available \href{https://github.com/Runjian-Chen/JiSAM}{here} 
\end{abstract}    
\section{Introduction}
\label{sec:intro}
Light-Detection-And-Ranging (LiDAR) provides accurate estimation of the distance between the sensor and objects and serves as one of the important sensors for autonomous driving, attracting broad research interests for 3D object detection with LiDAR \cite{pointpillars, pointrcnn, second, centerpoint, sst, transfusion, pv-rcnn, pv-rcnn++}. However, these learning-based LiDAR detectors heavily relies on labeled real-world LiDAR datasets, which hinders the real-world deployment of these methods. First of all, labeling in 3D space is notoriously \textbf{time-and-energy-consuming}. According to previous research works \cite{flava,nuscenes}, it typically costs an expert at least 10 minutes to label one frame of 3D data at a coarse level, which leads to more than 1000 days labeling for a one-hour sequence. Moreover, real world datasets are unable to cover all \textbf{corner cases} like rare traffic participants, which makes the trained models fail to detect such exceptional cases.

Previous research attempts on alleviating the cumbersome labeling of LiDAR point clouds can be divided into two main streams including semi-supervised learning \cite{pseudo-labeling, offboard, semi, semi-sampling,guided_point} and large-scale pre-training \cite{ad-pt, spot, gcc-3d, strl, co3, gd-mae, mv-jar, ponder, ponderv2, unipad}. But there exists two common problems of these methods. First of all, performance of these methods still lags far behind State-of-the-arts results when training on fewer real labels. Secondly, all these methods are unable to handle corner cases such as categories that are not labeled in the training set.

On the contrary, labeled LiDAR point clouds with corner cases can be easily obtained in modern simulator like CARLA \cite{CARLA} or Issac \cite{isaac}. Previous works including \cite{adversarial_sim_to_real, segmentation_sim_to_real, resimad} try to \textbf{directly} incorporate synthetic data from simulators to train models. However, there still exists large performance gap between models trained by these methods and models trained on real data. This observation shows that synthetic point clouds are still not ready to help 3D perception in real world application and it mainly stems from two challenges when introducing simulation point cloud into real-world 3D perception. 1) sample efficiency of simulation datasets is important for joint training on real and simulation datasets. As simulation data is less informative than real data, we might need far more simulation data. But the training and saving (disk capacity) cost increases as we scale up the simulation datasets. 2) simulation-to-real gap makes it hard to jointly train data in both domains (simulation and real scenes). For example, intensity of each point in CARLA simulator is computed directly via a function of xyz location while in real world, intensity reflects information about incident angle and surface material \cite{wang2021intensity}. Also, the 3D shapes of objects in the simulator differ from those in real world, resulting in various local point distribution for the same type of objects.

To overcome these challenges, we propose \textbf{Ji}ttering augmentation, domain-aware backbone and memory-based \textbf{S}ectorized \textbf{A}lign\textbf{M}ent Loss, namely \method\ . (1) jittering augmentation adds random noises to simulation LiDAR point clouds in Spherical Coordinate, inspired by previous literature \cite{lidar_noise_1,lidar_noise_2} where LiDAR noise is modelled as independent identical zero-mean gaussian distributions for different axis in Spherical Coordinate. This improves the sample efficiency of the simulation data. (2) domain-aware backbone utilizes different input kernels for both domains to adapt to various numbers of valid input channels. This makes use of all the available information in separate domains to improve the performance and increase the parameter number by less than $0.025\%$ only in training phase. (3) To further bridge the simulation-to-real gap, memory-based sectorized domain alignment loss is proposed. Memory-based sectorized alignment loss is based on the observation that if two objects of the same category have similar yaw rotation and appear in the same sector of the surrounding environments, the points distributions of these objects scanned by the LiDAR sensor are similar. An illustration of this observation is provided in Section 2 of Appendix. As shown in (c) in Figure \ref{figure: pipeline of proposed method}, we first divide the surrounding environment of the autonomous vehicle into different sectors and also divide headings into several bins. Then for each category in each sector, we create object-level feature memories for different heading bins. During training, we update these memories using features from real point clouds and align those of simulation point clouds to memories. These close the simulation-to-real domain gap and fully utilize simulation data to compensate the lack of real labeled data. To be noted, \method\ \ is an add-in module that can be applied to different 3D LiDAR detectors with few efforts.

To the best of our knowledge, \method\ \ is the first work able to train the SOTA 3D detector with only $2.5\%$ real data labeled and simulation data to achieve comparable performance. Besides, as the simulation-to-real gap is narrowed down, \method\ further improves corner case ability and objects not labeled in real data can be detected with the help of simulation data. Our findings open up the door to incorporate simulation data for real-world 3D LiDAR perception and we believe that it would close the gap between current DL-AD research community and real-world applications.
\section{Related Work}
\label{sec:related work}

\noindent\textbf{LiDAR 3D Object Detection. }Existing LiDAR-based 3D object detectors can be classified into three categories by the encoder used in the detectors. The first one is point-based methods and \cite{pointrcnn} is one of the representative work to use point-level embedding for 3D object detection. The second line is voxel-based methods embraced by \cite{second, centerpoint, transfusion, sst,voxelnext} which voxelize raw point clouds and use sparse 3D convolution to embed the 3D voxels. The third one is point-voxel-combination methods \cite{pv-rcnn, pv-rcnn++} that combine the voxel-level and point-level features for 3D embedding.

\paragraphskip
\noindent\textbf{Label-efficient LiDAR Perception. }As labeling for LiDAR point clouds is costly, there has been large research interest on reducing labeling burden for LiDAR perception and there are two main streams of work. (1) Embraced by pseudo-labeling \cite{pseudo-labeling} and its follow-up works \cite{offboard, semi, semi-sampling,guided_point}, the first branch is semi-supervised learning that utilizes both fewer labeled data and large amount of unlabeled data for training the 3D perception models. (2) The second way is to use large-scale pre-training and fine-tune the pre-trained backbones in different downstream datasets with fewer labels. AD-PT \cite{ad-pt} and SPOT \cite{spot} are representative works for semi-supervised pre-training for 3D detection on LiDAR point cloud and demonstrate strong performance gain when using fewer labels. Other works including GCC-3D \cite{gcc-3d}, STRL \cite{strl}, CO3 \cite{co3}, GD-MAE \cite{gd-mae} and MV-JAR \cite{mv-jar} utilize unlabeled data for pre-training and learned general features for downstream tasks. Nonetheless, when using less real LiDAR point clouds, all these works lags far behind the results trained with full training set. 

\paragraphskip
\noindent\textbf{Synthetic Point Clouds for Real Perception. } There exists some works \cite{resimad,adversarial_sim_to_real} in applying models trained in synthetic point clouds to real perception. However, performance of these methods still lags far behind State-of-the-arts results when training on fewer real labels, even when large-scale real-world labeled datasets are utilized. And \cite{resimad} is unable to handle corner cases such as categories that are not labeled in the training set.

\begin{figure*}[t]
\centering
\includegraphics[width=0.98\linewidth]{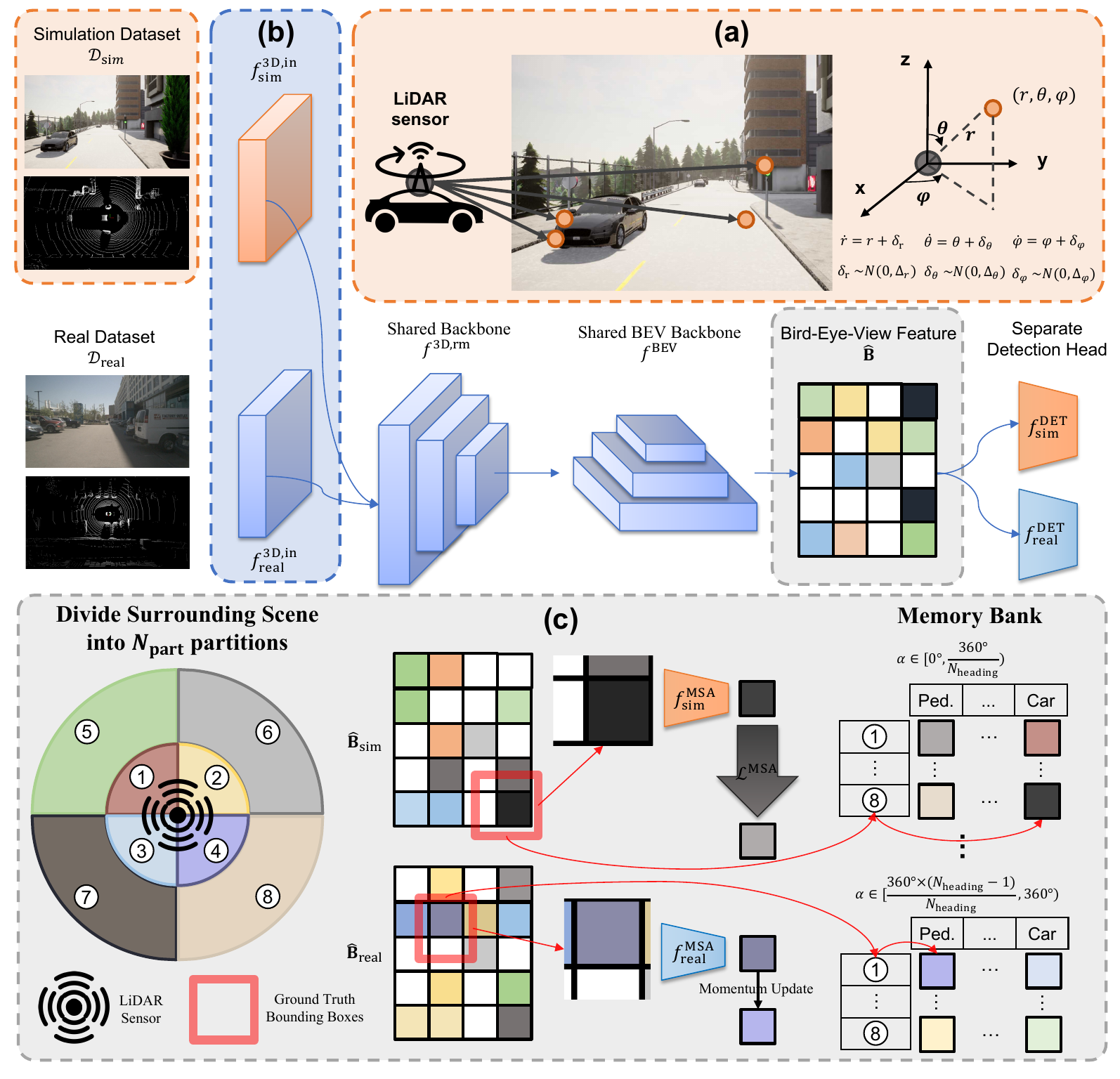} 
\vspace{-3mm}
\caption{The pipeline of the proposed method. Current SOTA LiDAR detectors are mainly consisted of three parts: 3D sparse backbone for embedding 3D voxels, BEV backbone for embedding bird-eye-view features and detection head to predict 3D bounding boxes from BEV features. \method\ \ jointly train simulation dataset from CARLA and a few labeled samples from real dataset. To increase sample efficiency and bridge the sim-to-real domain gap, we propose (a) jittering augmentation on noiseless simulation data, which largely increases the sample efficiency of simulation data and save the cost of training time and disk space (b) separate input embedding layer, which fully utilizes all useful information from both domain (c) memory-based sectorized alignment on BEV features to bridge the sim-to-real gap. This is inspired by the observation that in the same sector of the autonomous vehicle's neighborhood, two objects of the same category having similar heading would have similar points distribution in the LiDAR scan.
}
\vspace{-3mm}
 \label{figure: pipeline of proposed method}
\end{figure*}

\section{Method}
\label{sec: method}
In this section, we introduce the proposed method \method\ . As detailed in Figure \ref{figure: pipeline of proposed method}, there are three main components in \method\ . (a) Noise Augmentation is proposed for simulation data to improve sample efficiency. (b) Domain-aware Backbone is introduced to fully utilize useful information from different domains. (c) Sectorized Domain Alignment Loss is proposed to overcome the sim-to-real gap. To start with, we discuss the notations, background and the problem formulation in Section \ref{subsec: notations and problem formulation}. Then we respectively introduce Noise Augmentation, Domain-aware Backbone design, and the alignment losses in Section \ref{method: augmentation on simulation LiDAR datasets.}, \ref{method: backbone design} and \ref{method: alignment losses}.

\subsection{Notations and Problem Formulation}
\label{subsec: notations and problem formulation}
\textbf{Notations.} To begin with, we denote LiDAR point cloud $\mathbf{P}\in \mathbb{R}^{N\times (3+d)}$ as the concatenation of the $xyz$-coordinate $\mathbf{C}\in \R^{N\times 3}$ and the point features $\mathbf{F}\in \R^{N\times d}$, which leads to $\mathbf{P}=[\mathbf{C},\mathbf{F}]$. Here $N$ denotes the number of points (or voxels) and $d$ is the number of point feature channels, which varies among different datasets. For example, $d=2$ in NuScenes dataset \cite{nuscenes} representing the intensity and collected timestamp of each point. Moreover, as the point intensity from CARLA simulator is computed via a linear function of xyz-coordinate, there is no more useful information from this only feature in simulation dataset and $d=0$ for all the simulation datasets. Each LiDAR sample is paired with the bounding box labels $\mathbf{Y}\in\R^{M\times 10}$ for 3D object detection, where $M$ is the number of the surrounding objects and each 3D bounding box contains 10 scalars indicating category, xyz-location, xyz-size (length, width and height), orientation with respect to z-axis ($\alpha$) and speeds along x and y axis respectively. Thus each data sample $\mathbf{D}=[\mathbf{P},\mathbf{Y}]$ is consisted by the LiDAR point cloud and the labels. To indicate data samples from different domains, we use the subscripts `real' and `sim'. For instance, real-world and synthetic point clouds are represented by $\mathbf{P}_{\text{real}}=[\mathbf{C}_{\text{real}},\mathbf{F}_{\text{real}}]\in\R^{N_{\text{real}}\times (3+d)}$ and $\mathbf{P}_{\text{sim}}=[\mathbf{C}_{\text{sim}}]\in\R^{N_{\text{sim}}\times 3}$ respectively.

\paragraphskip
\noindent\textbf{Backgrounds. }We provide a brief review and summary for current SOTA LiDAR detectors \cite{transfusion,centerpoint,pv-rcnn++}. Firstly, a 3D sparse backbone $f^{\text{3D}}$ is applied for embedding raw LiDAR point clouds into 3D point/voxel-level features $\hat{\mathbf{P}}\in\mathbb{R}^{\hat{N}\times(3+\hat{d})}$. $f^{\text{3D}}$ is consisted of several 3D sparse convolution blocks including the input block $f^{\text{3D,in}}$ and the other remaining blocks $f^{\text{3D,rm}}$.
\begin{equation}
\hat{\mathbf{P}}=f^{\text{3D}}(\mathbf{P})=f^{\text{3D,rm}}(f^{\text{3D,in}}(\mathbf{P}))
\end{equation}
Here we use $\hat{N}$ to denote the number of points/voxels after encoding because pooling operation often exists in 3D encoders \cite{spconv,pointnet++} and changes the number of points/voxels. Moreover, $\hat{d}$ is the number of feature channels after encoding. Then the 3D features $\hat{\mathbf{P}}$ are mapped onto Bird-Eye-View features $\mathbf{B}\in\mathbb{R}^{H\times W\times\hat{d}}$, where $H$ and $W$ indicate the dimensions of the BEV features. The BEV features are fed into BEV backbone $f^{\text{BEV}}$ consisted of 2D convolution blocks.
\begin{equation}
\hat{\mathbf{B}}=f^{\text{BEV}}(\mathbf{B})
\end{equation}
where $\hat{\mathbf{B}}\in\mathbb{R}^{H\times W\times\dot{d}}$ is the embedded BEV features, where $\dot{d}$ is the number of feature channels after encoding. $f^{\text{BEV}}$ is a U-Net encoder and makes the spatial dimension of $\hat{\mathbf{B}}$ the same as that of $\mathbf{B}$. Finally a detection head $f^{\text{DET}}$ takes $\hat{\mathbf{B}}$ as inputs and make predictions for 3D bounding boxes $\hat{\mathbf{Y}}\in\R^{\hat{M}\times 10}$, with which we compute the detection loss $\mathcal{L}^{\text{DET}}(\hat{\mathbf{Y}}, \mathbf{Y})$ to optimize the network.
\begin{equation}
\hat{\mathbf{Y}}=f^{\text{DET}}(\hat{\mathbf{B}})
\end{equation}

\paragraphskip
\noindent\textbf{Problem Formulation. }For a real-world dataset $\mathcal{D}_{real}$, we first simulate a similar LiDAR configuration in the CARLA simulator and generate large amount of synthetic data with bounding box labels. Then we build a joint dataset $\mathcal{D}=\{\hat{\mathcal{D}}_{\text{real}},\mathcal{D}_{\text{sim}}\}$ with fewer real training samples $\hat{\mathcal{D}}_{\text{real}}=\{[\mathbf{P}_{\text{real}},\mathbf{Y}_{\text{real}}]\}$ and large amount of synthetic data $\mathcal{D}_{\text{sim}}=\{[\mathbf{P}_{\text{sim}},\mathbf{Y}_{\text{sim}}]\}$. In our experiments, $|\hat{\mathcal{D}}_{\text{real}}|\approx 7,000$ and $|\mathcal{D}_{\text{sim}}|\approx 80,000$, where $|\cdot|$ indicates the size of a set. Then we train current SOTA 3D detection models on $\mathcal{D}$ and tested on the whole validation set in $\mathcal{D}_{real}$.

There exists three main challenges to introduce simulation data for boosting model performance on real data. Firstly, although data generation in simulator is much easier and faster than real world data collection, the training cost increases when we scale up $\mathcal{D}_{\text{sim}}$. Thus sample efficiency of $\mathcal{D}_{\text{sim}}$ influences the training process. Second, as the number of useful point feature channels varies across datasets, how to fully utilize meaningful information in different datasets remains an open question. Last but not least, the sim-to-real gap hinders the performance improvement when jointly training models on $\mathcal{D}$. To overcome these three challenges, we propose Noise Augmentation for $\mathbf{P}_{sim}$, Domain-aware Backbone with shared weights except for separate input layers $f^{\text{3D,in}}_{real/sim}$ for different domains and Sectorized Domain Alignment Loss on BEV features, $\mathcal{L}^{\text{SMA}}$. Thus the final loss function is consisted of three parts, where $\lambda$ are weights to balance the loss function.
\begin{equation}
\mathcal{L}=\mathcal{L}^{\text{DET}}(\hat{\mathbf{Y}}, \mathbf{Y}) + \lambda\mathcal{L}^{\text{SMA}}(\hat{\mathbf{B}})
\end{equation}

\subsection{Noise Augmentation}
\label{method: augmentation on simulation LiDAR datasets.}
Although countless simulation LiDAR data can be generated inside the CARLA simulator, the training cost grows rapidly with the increase in data volume and it is infeasible to store too much data with limited resources. To improve sample efficiency, we propose Noise Augmentation for the simulation dataset. We start with discussion on the point cloud generation process for LiDAR and introduce how Noise Augmentation works.

\paragraphskip
\noindent\textbf{LiDAR Point Clouds Generation. }As shown in Figure \ref{figure: pipeline of proposed method} (a), LiDAR emits laser beams to the surrounding environment and then laser beams hit objects/background and get reflected back towards the LiDAR. By measuring the time difference between emission and detection of the laser beams, the LiDAR system can calculate the distance between the sensor coordinate and the detected surface (with the known speed of light), that is $r$ in Figure \ref{figure: pipeline of proposed method} (a). Together with the built in emission angel $\theta$ and $\varphi$, we get the point clouds in Spherical coordinate system. Noise exists during the sensing process and according to the statistic model in previous literature \cite{lidar_noise_1,lidar_noise_2}, LiDAR noise can be modelled as independent identical zero-mean gaussian distributions for $r$, $\theta$ and $\varphi$. Thus the LiDAR point cloud generation process can be formulated as belows, where $\Delta_{r}$, $\Delta_{\theta}$ and $\Delta_{\varphi}$ respectively represent variance of the noise on $r$, $\theta$ and $\varphi$.
\begin{equation}
\label{equation: noise model in LiDAR point clouds}
\left\{
\begin{aligned}
   \dot{r} &=  r + \delta_{r}, &\delta_{r} \sim N(0,\Delta_{r})\\
   \dot{\theta} &= \theta + \delta_{\theta}, &\delta_{\theta} \sim N(0,\Delta_{\theta})\\
   \dot{\varphi} &=  \varphi + \delta_{\varphi}, &\delta_{\varphi} \sim N(0,\Delta_{\varphi})
\end{aligned}
\right.
\end{equation}

To further compute the xyz coordinate for each point, transformation between Spherical and Cartesian coordinates is applied and we arrive at the $\mathbf{P}$ in Cartesian coordinates.
\begin{equation}
\label{equation: spherical to cartesian}
\left\{
\begin{aligned}
   x &= \dot{r} \sin{\dot{\theta}}\cos{\dot{\varphi}}\\
   y &= \dot{r} \sin{\dot{\theta}}\sin{\dot{\varphi}}\\
   z &= \dot{r} \cos{\dot{\theta}}
\end{aligned}
\right.
\end{equation}

\paragraphskip
\noindent\textbf{Noise Augmentation. }The noise level ($\Delta_{r}$, $\Delta_{\theta}$ and $\Delta_{\varphi}$)  can be controlled in the CARLA simulator. To improve sample efficiency of simulation dataset, we set zero variance when collecting the simulation data and apply Noise Augmentation during training, which significantly increases data diversity with the same amount of simulation data. Firstly, in each training iteration, if the data sample is from simulation dataset, we transform the LiDAR point cloud $\mathbf{P}_{\text{sim}}$ back to the Spherical coordinate.
\begin{equation}
\left\{
\begin{aligned}
   r &= \sqrt{x^2+y^2+z^2}\\
   \theta &= \arccos{\frac{z}{r}} \\
   \varphi &= \operatorname{sgn}(y)\arccos{\frac{x}{\sqrt{x^2+y^2}}}
\end{aligned}
\right. \ \ ,\ \ \operatorname{sgn}(y)=
\left\{
\begin{aligned}
-1\ \ &\text{if}\ y < 0, \\
0\ \ \ &\text{if}\ y = 0, \\
1\ \ \ &\text{if}\ y > 0.
\end{aligned}
\right.
\end{equation}
Then we draw random noise $\delta_{r/\theta/\varphi}$ from zero-mean gaussian distribution with $\Delta_{r/\theta/\varphi}$. And following equation \ref{equation: noise model in LiDAR point clouds} and \ref{equation: spherical to cartesian}, we add the random noise to $\mathbf{P}_{\text{sim}}$ and transform it back to the Cartesian coordinate.
The random noise is different for the same simulation data sample in different epochs, making the surface point distribution (local distribution) for different objects more diverse with the same amount of simulation data.

\subsection{Domain-aware backbone}
\label{method: backbone design}
The number of point feature channels $d$ varies across simulation and real domains. For simulation datasets, we only use xyz-location and timestamp because intensity in CARLA simulator is directly computed via a linear function of xyz location, making it meaningless. For data samples in NuScenes \cite{nuscenes}, each point contain intensity and timestamp, both of which are useful for object detection. To make full use of all informative input channels, we apply separate input blocks $f^{\text{3D,in}}_{\text{sim/real}}$ for different domains, which only brings less than $0.025\%$ computational overhead during training. All the remaining blocks in the 3D backbone are shared.

\subsection{Alignment Losses}
\label{method: alignment losses}
Another important sim-to-real gap lies in the difference between 3D shape of synthetic assets and that of real-world objects. In the coordinate of the autonomous vehicle, it can be observed that for objects in the same class, which have similar headings and appear in the same sector in the neighborhood, the point distributions scanned by the LiDAR sensor appear to be quite similar. We encourage readers who are interested in obervation to take a look at Figure 2 in the Appendix, which is an illustration of this observation. To further bridge the sim-to-real gap, we propose memory-based sectorized alignment loss. First of all, we introduce how we build up the memory bank. Then we show details on how to update the memory and use the memory for loss computation.



\paragraphskip
\noindent\textbf{Build up Memory Bank. }We divide the surrounding environment of the autonomous vehicle into $N_{\text{sc}}$ using shape context \cite{shape_context_1, shape_context_2, shape_context_3, shape_context_4}. Figure \ref{figure: pipeline of proposed method} (c) shows an example when $N_{\text{sc}}=8$. Although point density of objects varies across different ranges, the distances from ego-vehicle to objects in the same partition are similar, which improves consistency of the memory bank. For the heading of objects (orientation against the z axis in ego-vehicle’s coordinate), we discretize 360 degrees into $N_{\text{heading}}$ bins. Finally, assuming there are $N_{\text{cls}}$ categories in the dataset, we create a memory bank $\textbf{M}\in\R^{N_{\text{sc}}\times N_{\text{heading}} \times N_{\text{cls}} \times \dot{d}}$ with random initialization, where $\dot{d}$ is the number of feature channels of the embedded BEV features $\hat{\mathbf{B}}$.

\paragraphskip
\noindent\textbf{Update Memory Bank. }With the ground truth labels of real objects $\mathbf{Y}_{real}$ and their embedded BEV features $\hat{\mathbf{B}}_{real}$, we extract the features of different objects, find their corresponding memory index and update the memory bank. Note that in order to avoid complex notation here, we eliminate the subscripts in the following equations. To extract the features of objects $\mathbf{O}\in\R^{M\times \dot{d}}$, we apply the RoI-grid pooling $f^{\text{RoI-pool}}$ \cite{pv-rcnn, pv-rcnn++}, which is a popular module for embedding local features.
\begin{equation}
\mathbf{O}=f^{\text{RoI-pool}}(\hat{\mathbf{B}}, \mathbf{Y})
\end{equation}
Then we compute the memory index $(n_{\text{sc}}, n_{\text{heading}}, n_{\text{cls}})$ for each object with its location, heading and category. Assume we are working on the $i^{th}$ object. (The process can be done in parallel and here is just a simple illustration) For category index, we directly map the category to an integer $n_{\text{cls}}$. For sector $n_{\text{sc}}$ and heading $ n_{\text{heading}}$ indexes of the $i^{th}$ object, we have
\begin{equation}
n_{\text{sc}} = f^{\text{SC}}(\mathbf{Y}_{i})\ ;\ n_{\text{heading}} = \lfloor \mathbf{H}_{i} \div \frac{360^{\circ}}{N_\text{heading}} \rfloor
\end{equation}
where $f^{\text{SC}}$ is the shape context operation \cite{shape_context_1, shape_context_2, shape_context_3, shape_context_4}. $\mathbf{Y}_{i}$ indicates the labels for the $i^{th}$ object and  $\mathbf{H}_{i}$ is headings of the $i^{th}$ object in its labels $\mathbf{Y}_{i}$. We then update the corresponding memory with momentum $m$ and the feature of the $i^{th}$ object $\mathbf{O}_{i}$.
\begin{equation}
\mathbf{M}_{n_{\text{sc}}, n_{\text{heading}}, n_{\text{cls}}, :} = m\times \mathbf{M}_{n_{\text{sc}}, n_{\text{heading}}, n_{\text{cls}}, :} + (1-m) \times \mathbf{O}_{i,:}
\end{equation}

\begin{figure*}[t]
\centering
\includegraphics[width=0.98\linewidth]{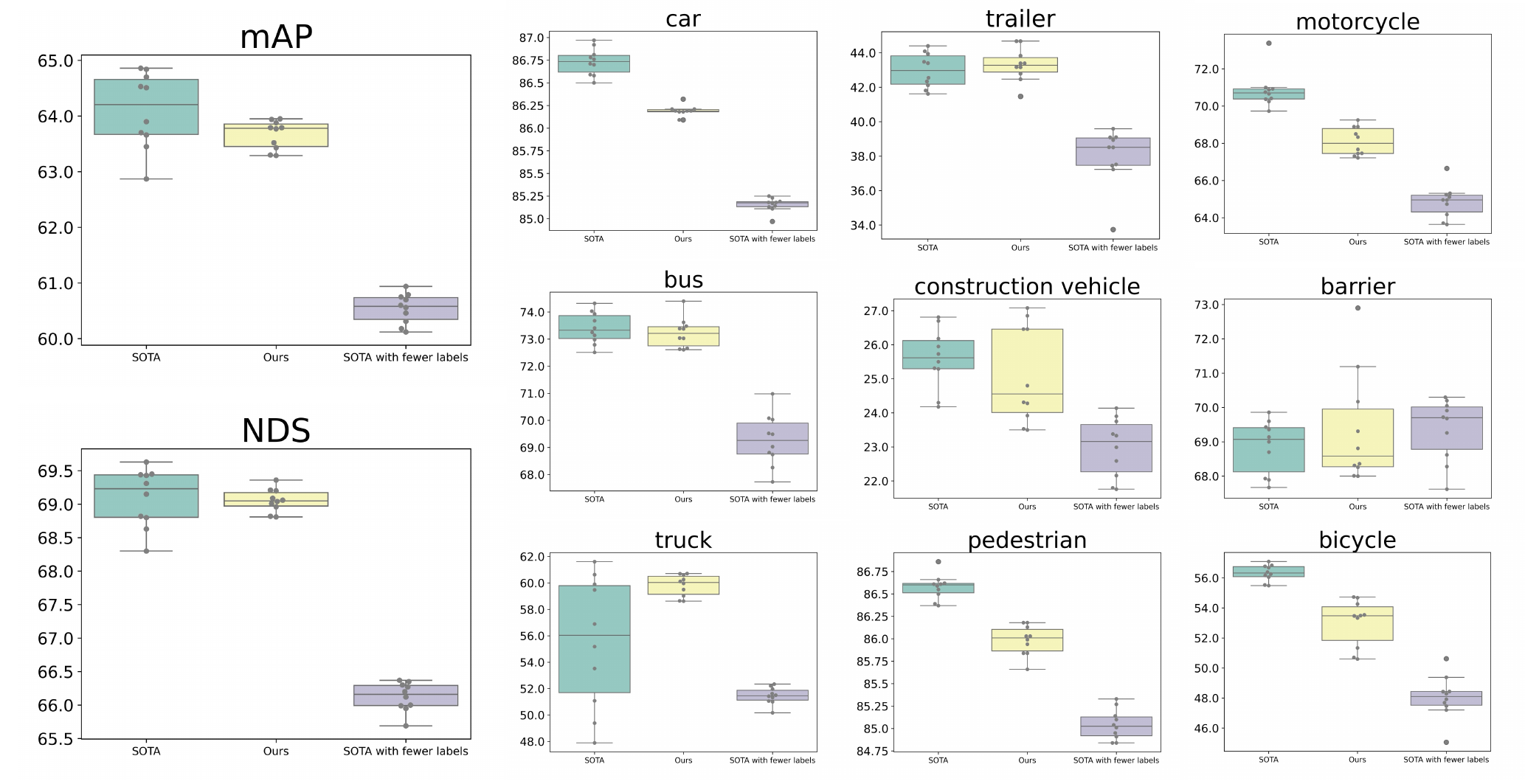} 
\vspace{-3mm}
\caption{Overall results on NuScenes Dataset. The SOTA 3D detector, Transfusion \cite{transfusion} is used for all the experiments. In the figure, `SOTA' means Transfusion trained on all the available labels in NuScenes dataset. `SOTA with fewer labels' means Transfusion trained on 2.5\% of all the LiDAR point clouds in the training set. 4 mAP and 3 NDS drops are observed. `Ours' means \method\ \ that utilizes only 7,000 labeled real LiDAR point clouds and simulation point clouds to train Transfusion. It can be found that \method\ improves the performance of `SOTA with fewer labels' by a significant margin and achieves comparable performance with Transfusion trained on all the available labels in NuScenes dataset.
}
\vspace{-3mm}
 \label{figure: overall results}
\end{figure*}

\paragraphskip
\noindent\textbf{Memory-based Sectorized Alignment Loss. }As in the beginning of training, both features from real and simulation domains contain little useful information, we set a warm-up stage for this loss, update the memory and apply this loss only after $N_{\text{warmup}}$ training iterations. During the warm-up process, the memories of less frequent objects are also updated, alleviating the influence of less frequent objects. After warming-up, we extract object features $\mathbf{O}$ for simulation data ($\mathbf{Y}_{sim}$, $\hat{\mathbf{B}}_{sim}$) and compute memory index $(n_{\text{sc}}, n_{\text{heading}}, n_{\text{cls}})$ for each of them using the same method as described above. Then a mean-squared-error loss function is applied on the simulation object features and their corresponding features in the memory.
\begin{equation}
\mathcal{L}^{\text{SMA}} = \frac{1}{M\times \dot{d}}\sum_{i=1}^{M}\sum_{j=1}^{\dot{d}}(\mathbf{O}_{i,j} - \mathbf{M}_{n_{\text{sc}}, n_{\text{heading}}, n_{\text{cls}}, j})^2
\end{equation}
In practice, we find that conducting the memory-based sectorized alignment loss in a bi-directional way introduce more improvement, which means that we separately maintain memory banks for both real $\mathbf{M}_{\text{real}}$ and simulation objects $\mathbf{M}_{\text{sim}}$ and compute the alignment loss with $<\mathbf{M}_{\text{real}}, \mathbf{O}_{\text{sim}}>$ and $<\mathbf{M}_{\text{sim}}, \mathbf{O}_{\text{real}}>$ pairs. Note that \method\ does not construct real memory for categories not in real data but if they exists in synthetic data, the simulation bank is created. Categories available in the memory banks are involved in computing $\mathcal{L}^{\text{SMA}}$.
\section{Experiments}
\label{sec: exps}
In this section, we design experiments to answer two question: 1) Could synthetic LiDAR point clouds help real-world 3D perception with the help of \method ? 2) Is \method\ able to help corner cases like unlabeled categories? We starts with a discussion on experiment setup in Section \ref{subsec: experiment settings}. Then main results are presented in Section \ref{subsec: main results}. Finally we provide a discussion in Section \ref{subsec: discussion}.

\begin{figure*}[t]
\centering
\includegraphics[width=0.98\linewidth]{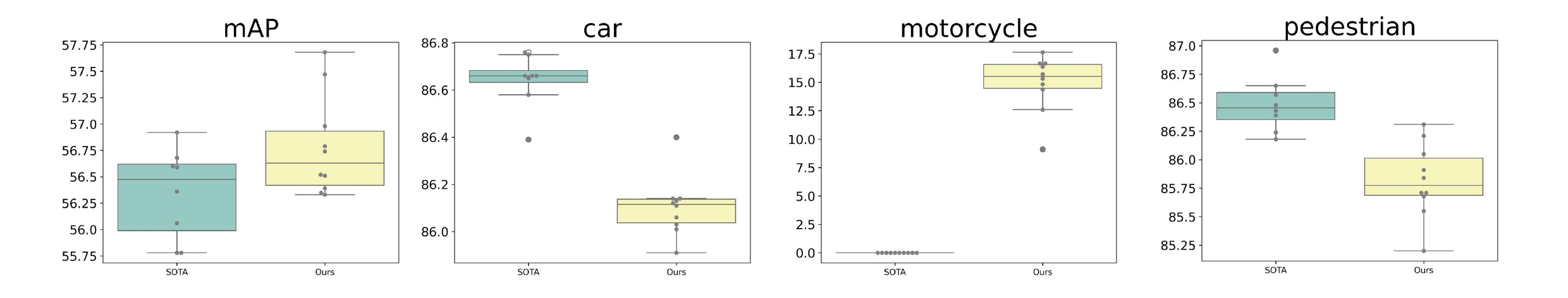} 
\vspace{-3mm}
\caption{Results on corner cases study. We manually eliminate the labels of motorcycle in the real training set to simulate the scene where corner labels (motorcycle) only exists in the evaluation set. The SOTA 3D detector, Transfusion \cite{transfusion} is used for all the experiments. In the figure, `SOTA' means Transfusion trained on all the available labels in NuScenes dataset and `Ours' indicates training Transfusion with \method\ with simulation data and 2.5\% of all the LiDAR point clouds in the training set. We present results of mAP and APs for some main categories. It can be found that even without labels of motorcycle in real dataset, \method\ achieves approximately 16\% mAP on this category on evaluation dataset and a bit better overall mAP. Meanwhile, for other categories like car and pedestrian, the performance are comparable (difference lower than 0.5\% APs).
}
\vspace{-3mm}
 \label{figure: corner case results}
\end{figure*}

\subsection{Experiment Settings}
\label{subsec: experiment settings}
\noindent\textbf{Dataset.} We conduct main experiments on the popular autonomous driving dataset NuScenes \cite{nuscenes}. The autonomous vehicle in NuScenes \cite{nuscenes} utilizes a 32-beam LiDAR (20Hz) to collect 20-second periods in 850 scenes, among which 700 scenes are selected as training scenes and the others for evaluation. Thus in the training set, there are approximately 280,000 LiDAR point clouds. The data is collected in Boston and Singapore for 15 hours' driving. We select the open source simulator CARLA \cite{CARLA} to create synthetic point clouds. We set up the same setting for LiDAR sensor and the height of autonomous vehicle. Then we collect simulation datasets in the scenarios provided in CARLA. To reduce disk storage, we reduce the frequency in simulation data collection. And this results in a dataset consisted of 80,000 labeled LiDAR point clouds with a larger data diversity than NuScenes \cite{nuscenes}. Dataset statistic is shown in Table 1 in Section 3 of Appendix.

\paragraphskip
\noindent\textbf{Baselines Method.} We utilize the current SOTA LiDAR 3D object detector Transfusion \cite{transfusion} as our baseline and use the implementation in the popular OpenPCDet \cite{openpcdet2020} repository.

\paragraphskip
\noindent\textbf{Main Experiments.} The goal of the experiments is to answer two questions: 1) Can \method\ \ introduce comparable SOTA performance with fewer real labels in order to reduce labeling burden? 2) How well is \method\ \ in handling corner cases? Firstly, we utilize $2.5\%$ of all the LiDAR point clouds in the training set of NuScenes, that is approximately 7,000 labeled LiDAR point clouds and the whole simulation dataset to train Transfusion with \method\ . Then for the second question, we manually eliminate some class (motorcycle) in real data and train Transfusion with it. This simulates the scenes when the 3D perception model encounters corner cases in the validation sets. Meanwhile, we use the joint dataset with $2.5\%$ of the training LiDAR point clouds in NuScenes without motorcycle labels and the whole simulation dataset to train Transfusion with \method\ \ to see how \method\ \ deal with corner cases.


\paragraphskip
\noindent\textbf{Results Presentation.} We use the common metric for NuScenes \cite{nuscenes} for evaluation, mean Average Precision (mAP), Nuscenes Detection Score (NDS) and Average Precision for each category. For main experiments, we randomly conduct the training for 10 times with 10 different random seeds. Then we visualize the results using box-plot \cite{boxplot}, which shows the first, median, third as well as the minimum and maximum quartiles of the results.

\paragraphskip
\noindent\textbf{Implementation Details of \method. }For loss weighting, we set $\lambda = 0.1$ and balance the loss from different domain by $\omega_{\text{real}}=1.0$ and $\omega_{\text{sim}}=0.1$. The deviation of noise level for Jittering Augmentation is set as $\Delta_{r}=0.01$ and $\Delta_{\theta/\varphi}=0.0001$. For Sectorized Domain Alignment loss, we set the number of partitions for shape context as $N_{\text{sc}}=32$ and divide headings into $N_{\text{heading}}=32$ bins. We use AdamW optimizer with one-cycle learning-rate schedule, where the peak learning rate is set as $Lr=0.004$. Weight decay is set to $0.01$. And we train \method\ \ for $100$ epochs. All the experiments are conducted on 4 A100 GPUs.

\subsection{Results and Analysis}
\label{subsec: main results}
\noindent\textbf{Reduce Labeling Burden.} In this part, we use \method\ \ to train Transfusion with approximately 7,000 labeled LiDAR point clouds and the whole simulation dataset and compare its performance with Transfusion trained with all the available labels in NuScenes dataset. As shown in Figure \ref{figure: overall results}, for overall performance evaluation including mAP and NDS (nuscenes detection scores), \method\ \ achieves comparable performance with Transfusion with fewer real labels. Also, due to the large diversity and sample number of simulation data, the variance of the performance of \method\ \ is smaller than Transfusion trained with all the real labels. Meanwhile, when compared to Transfusion trained on same number of real labels, \method\ \ achieves significant performance gain of approximately 4 mAP and 3 NDS. When we look into more detailed results for different categories, \method\ \ achieves even better Average Precision than Transfusion trained with all the available labels in truck. This is because in real data, the number of trucks is much smaller than cars but in CARLA, we can create numerous trucks to help the detection of real trucks. It can be found that compared to `SOTA with fewer labels', the performance gain in truck is larger than 8 AP. For other categories like car, trailer, bus, construction vehicle, barrier and pedestrian, \method\ \ achieves comparable performance as that of Transfusion trained with 100\% labels. And for these categories and motorcycle, bicycle, the performance gains over Transfusion trained with the same number of real labels, are always larger than 1 AP. This is because in simulator, we can generate numerous objects in these categories to compensate the lack in real labels. For some categories like motocycle and bicycle, the performance of \method\ \ is a bit worse than Transfusion trained on full labels. This might stem from two reasons: 1) The number of available 3D models of motorcycle and bicycle in CARLA is small, making the diversity of simulated motorcycles and bicycles a bit smaller than the whole real dataset.  2) The scanned points of this two categories are scarse, which further makes it harder for the network to learn to generalize to real world objects.

\begin{table}[t]
\centering
\begin{tabular}{ccccc}
\hline
Model           & Real & Sim & mAP   & NDS   \\ \hline
Tranfusion \cite{transfusion}    &            \ding{51}          &     \ding{55}                 & 64.51 & 69.31 \\
Tranfusion with \method &       \ding{55}               &            \ding{51}          & 63.95 & 69.36 \\
VoxelNext \cite{voxelnext}    &            \ding{51}          &     \ding{55}                 & 60.53 & 66.65 \\
CenterPoint \cite{centerpoint}   &         \ding{51}             &    \ding{55}                  & 59.22 & 66.48 \\
Second \cite{second}     &          \ding{51}            &      \ding{55}                & 50.59 & 62.29 \\
PointPillar \cite{pointpillars}    &          \ding{51}            &       \ding{55}               & 44.63 & 58.23 \\ \hline
\end{tabular}
\vspace{-1mm}
\caption{Comparison to different 3D object detectors. ``Real'' means whether the model is trained on all real labels in NuScenes. ``Sim'' indicates whether the model is trained with simulation data.}
\label{table: comparison to 3d detectors}
\end{table}

\paragraphskip
\noindent\textbf{Corner Case Results.} In this part, we manually eliminate all motorcycle labels in the training set of NuScenes \cite{nuscenes} and train Transfusion \cite{transfusion} with all the available data, which is indicated by `SOTA'. Then, we use \method\ \ to jointly train Transfusion \cite{transfusion} with approximately 7,000 LiDAR point clouds in the real training set and the whole simulation dataset. The results of the two training is shown in Figure \ref{figure: corner case results}. 
We present the overall mAP and APs of some main categories. It can be found that although \method\ \ uses much fewer real labels, it achieves better performance than `SOTA' in overall mAP (up to 1\% mAP improvement). For missing category (motorcycle) in training set, \method\ \ achieves around 16\% APs, which can significantly guarantee safety for corner cases. For other categories like car and pedestrian, the performance is comparable with a difference within 0.5\% APs even when \method\ \ is trained by much fewer real labels.


\paragraphskip
\noindent\textbf{Comparison to Other 3D Detectors. }We provide the results comparing Transfusion with \method\ \ to other popular 3D object detectors on NuScene in Table \ref{table: comparison to 3d detectors}. It can be found that \method\ \ is able to utilize simulation data and much fewer real labels to achieve SOTA performance.

\begin{table}[t]
\centering
\begin{footnotesize}

\begin{tabular}{cccccc}
\hline
Syn. & D.S.B. & S.A. & J.A. & mAP   & NDS   \\ \hline
      
      \ding{55}          & 
                \ding{55}         &            \ding{55}            &      \ding{55}                             & 60.18	 & 65.69 \\

 \ding{52}          & 
                \ding{55}         &            \ding{55}            &      \ding{55}                             &  58.29 (\textcolor{BrickRed}{-1.89})	 & 64.71 (\textcolor{BrickRed}{-0.98}) \\

\ding{51}          & 
\ding{51}     & \ding{55}   & \ding{55}             & 62.30 (\textcolor{forest}{+2.12}) & 68.07 (\textcolor{forest}{+2.38}) \\

\ding{51}          & 
 \ding{51}     & \ding{51}   & \ding{55}           & 63.71 (\textcolor{forest}{+3.53}) & 69.09 (\textcolor{forest}{+3.40})  \\

\ding{51}          &           \ding{51}            &       \ding{55}                 &      \ding{51}                             & 63.28 (\textcolor{forest}{+3.10})  & 68.89 (\textcolor{forest}{+3.20})  \\

\ding{51} ($50\%$)         &          \ding{51}          &              \ding{51}          &      \ding{51}                             & 63.90 (\textcolor{forest}{+3.72})  & 68.97 (\textcolor{forest}{+3.28})  \\
                
\ding{51}          &          \ding{51}          &              \ding{51}          &      \ding{51}                             & 63.95 (\textcolor{forest}{+3.77})  & 69.36 (\textcolor{forest}{+3.67})  \\

                \hline
\end{tabular}
    
\end{footnotesize}
\vspace{-1mm}
\caption{Ablation study results. ``Syn.'' describes whether synthetic data is used in the training. ``D.S.B.'' refers to Domain-specific backbone. ``S.A'' is the Memory-based Sectorized Alignment Loss. ``J.A.'' means Jittering Augmentation. It can be found that simply add synthetic data into training degrade the performance and each component proposed in \method\ \ brings performance improvement.}
\vspace{-2mm}
\label{table: ablation study}
\end{table}

\paragraphskip
\noindent\textbf{Ablation Study. }To verify the effectiveness of each proposed component, we design ablation experiments. More specifically, we mitigate one of the three components in \method\ \ each time and train Transfusion with it on the joint dataset with $2.5\%$ of the training LiDAR point clouds in NuScenes and the whole simulation dataset. We use the fix random seed in OpenPCDet \cite{openpcdet2020} to conduct the experiments, which makes the results \textit{reproducible}. In Table \ref{table: ablation study}, we present the quantitative results of ablation study. When comparing the last two rows, it can be found that simply add synthetic data into training even degrades the overall performance in mAP and NDS. And we consequently add domain-specific backbone, memory-based sectorized alignment loss and jittering augmentation and results show that each component bring their own benefits, which demonstrates the effectiveness of each component in \method\ . And if we compare the fourth and sixth lines, it can be found that with $50\%$ synthetic data and jittering augmentation, performance is comparable to all synthetic data without jittering augmentation, which demonstrates that jittering augmentation can help increase sample efficiency of synthetic data.

\subsection{Discussion}
\label{subsec: discussion}

\noindent\textbf{Seamless Integration with Other 3D Detectors.} As \method\ \ is a plug-and-play module, it can be seamlessly integrated with other 3D detectors like CenterPoint \cite{centerpoint} with minimal adjustment, which makes it general to different datasets and detection architecture. 

\paragraphskip
\noindent\textbf{Orthogonal to State-of-The-Art LiDAR Generative Model.} \method\ explore how to efficiently utilize simulation dataset and bridge the simulation-to-real gap with synthetic data generated from simulation engine, which relies on no real datasets. On the contrary, current State-of-The-Art LiDAR generative models \cite{lidar_generative_model_1, lidar_generative_model_2, lidar_generative_model_3, lidar_generative_model_4} are trained on large-scale real datasets and some methods even require all the labels. One important thing is that \method\ \ is orthogonal to all these generative models and performance might be further improved when we use LiDAR point clouds generated by them as simulation datasets. Besides, as \method\ is able to bridge the simulation-to-real and achieve comparable SOTA performance, it can also be used as a generation-quality testing tools to the LiDAR generative models.
\section{Conclusion}
\label{sec: conclusion}
In this paper, we propose \method\ to utilize the simulation LiDAR point clouds from modern simulators like CARLA \cite{CARLA} to help real 3D perception. \method\ is able to utilize a dataset , consisted of labels on only 2.5\% of all available real point clouds and together with large amount of simulation data, to achieve comparable performance with SOTA 3D object detector trained on the whole real training set. We believe this work can not only facilitate understanding of LiDAR perception but also bring the AD community one step closer to real-world deployment.
\section*{Acknowledgments}
This paper is partially supported by the General Research Fund of Hong Kong No. 17200622 and 17209324.
{
    \small
    \bibliographystyle{ieeenat_fullname}
    \bibliography{main}
}


\end{document}